\documentclass{article}

\usepackage{nonatbib}
\usepackagewithoutnatbib{jmlr2e}
\usepackage{lastpage}
\usepackage[frozencache,cachedir=minted]{minted}
\usepackage{subcaption}
\usepackage{tikz}
\usepackage{xcolor}
\definecolor{node_yellow}{HTML}{FDE725}

\usepackage[authoryear,citations,centerfigures,commands,enumerate,environments]{AVT}

\usepackage[capitalize]{cleveref}
\bibliography{references}

\title{The GeometricKernels Package: Heat and Matérn Kernels for Geometric Learning on Manifolds, Meshes, and Graphs}

\author{\name Peter Mostowsky\textsuperscript{\ensuremath{\dagger}}
\email pmostowsky@gmail.com \\
\addr Independent
\AND
\name Vincent Dutordoir 
\email vd309@cam.ac.uk \\
\addr University of Cambridge
\AND
\name Iskander Azangulov 
\email iskander.azangulov@spc.ox.ac.uk \\
\addr University of Oxford
\AND
\name Noémie Jaquier 
\email jaquier@kth.se\\
\addr KTH Royal Institute of Technology
\AND
\name Michael John Hutchinson 
\email michael.hutchinson@univ.ox.ac.uk \\
\addr University of Oxford
\AND
\name Aditya Ravuri
\email ar847@cam.ac.uk \\
\addr University of Cambridge
\AND
\name Leonel Rozo
\email leonel.rozo@ai4i.it \\
\addr Italian Institute of Artificial Intelligence for Industry
\AND
\name Alexander Terenin\textsuperscript{\ensuremath{\dagger}} 
\email avt28@cornell.edu \\
\addr Cornell University
\AND
\name Viacheslav Borovitskiy\textsuperscript{\ensuremath{\dagger}}
\email viacheslav.borovitskiy@gmail.com \\
\addr ETH Zürich and University of Edinburgh
}

\newcommand{\fixwidth}[1]{\ifdefined\XeTeXversion#1\else\textls[-3]{#1}\fi}

\jmlrheading{26}{2025}{1-\pageref{LastPage}}{7/24; Revised
11/25}{12/25}{24-1185}{Peter Mostowsky, Vincent Dutordoir, Iskander Azangulov, Noémie Jaquier, Michael John Hutchinson, Aditya Ravuri, Leonel Rozo, Alexander Terenin, and Viacheslav Borovitskiy}
\ShortHeadings{The GeometricKernels Package: Heat and Matérn Kernels for Geometric Learning}{\fixwidth{Mostowsky, Dutordoir, Azangulov, Jaquier, Hutchinson, Ravuri, Rozo, Terenin, Borovitskiy}}

\editor{Zeyi Wen}

\begin{document}

\maketitle

\begin{abstract}
Kernels are a fundamental technical primitive in machine learning. In recent years, kernel-based methods such as Gaussian processes are becoming increasingly important in applications where quantifying uncertainty is of key interest. In settings that involve structured data defined on graphs, meshes, manifolds, or other related spaces, defining kernels with good uncertainty-quantification behavior, and computing their value numerically, is less straightforward than in the Euclidean setting. To address this difficulty, we present \textsc{GeometricKernels}, a Python software package which implements the geometric analogs of classical Euclidean squared exponential---also known as heat---and Matérn kernels, which are widely-used in settings where uncertainty is of key interest. As a byproduct, we obtain the ability to compute Fourier-feature-type expansions, which are widely used in their own right, on a wide set of geometric spaces. Our implementation supports automatic differentiation in every major current framework simultaneously via a backend-agnostic design. In this companion paper to the package and its documentation, we outline the capabilities of the package and present an illustrated example of its interface. We also include a brief overview of the theory the package is built upon and provide some historic context in the appendix.
\end{abstract}

\begin{table}[b!]
\raggedright
\footnoterule
\footnotesize\textsuperscript{\ensuremath{\dagger}}Corresponding authors.
\\
Website: \url{https://geometric-kernels.github.io}.
\\
Code available at: \url{https://github.com/geometric-kernels/GeometricKernels}.
\end{table}

\section{Introduction}

Kernel methods, and methods built on them such as Gaussian processes, are a fundamental part of the modern machine learning toolkit.
The growth of machine learning has led to models being adopted in \emph{geometric} settings that involve structured data defined over spaces such as graphs, meshes, and their smooth analogs such as Riemannian manifolds.
An increasing need for kernel methods in such settings---particularly those with principled and intuitive handling of uncertainty, such as symmetry-based kernels---has led to a recent growth in research on \emph{geometric kernels} and their applications in fields such as robotics and neuroscience: examples are given in \Cref{sec:history}.

A key challenge in geometric settings is that kernels, to be well-defined, need to be positive semi-definite functions.
In geometric settings---in contrast to the classical Euclidean case---it is generally unclear how to define such functions directly using closed-form analytic expressions.
Straightforward generalizations, such as the distance-based squared exponential kernel, fail to be positive semi-definite in most cases \cite{schoenberg1938, feragen2015, dacosta2023}.
As a result, many kernels are implicitly-defined---so, their numerical computation tends to be more involved than of Euclidean kernels.
Our key motivation is to develop software that handles this difficulty well.

We present \textsc{GeometricKernels}, a software package designed to make geometric kernels as easy to work with in practice as possible.
We provide implementations of the geometric analogs of classical Euclidean squared exponential and Matérn kernel classes, which can be used within kernel learning pipelines in a largely plug-and-play manner.
We support graphs, meshes, and a collection of mathematically tractable manifolds.
Our implementation is designed to seamlessly support GPUs and automatic differentiation using all major frameworks---\textsc{PyTorch}, \textsc{JAX}, and \textsc{TensorFlow}---simultaneously. 
We also support \textsc{Numpy} to facilitate debugging and other use cases.

\section{The GeometricKernels Package}
\label{sec:software}
\begin{figure}
\begin{tikzpicture}
\node at (0,0) {\includegraphics[scale=0.25]{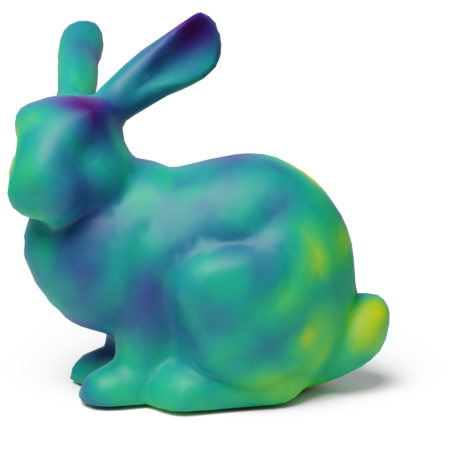}};
\node at (3.25,0) {\includegraphics[scale=0.25]{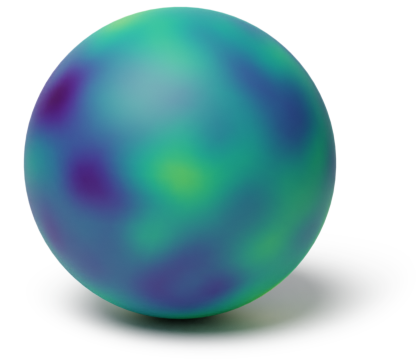}};
\node at (7.25,0) {\includegraphics[scale=0.25]{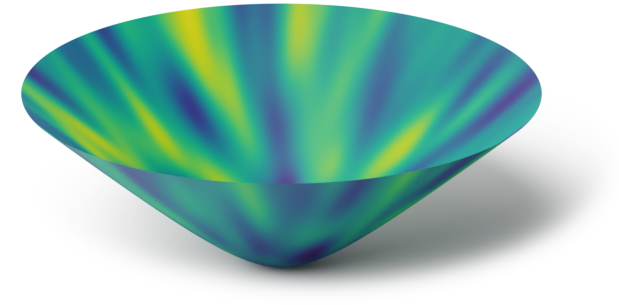}};
\node at (11.25,0) {\includegraphics[scale=0.25]{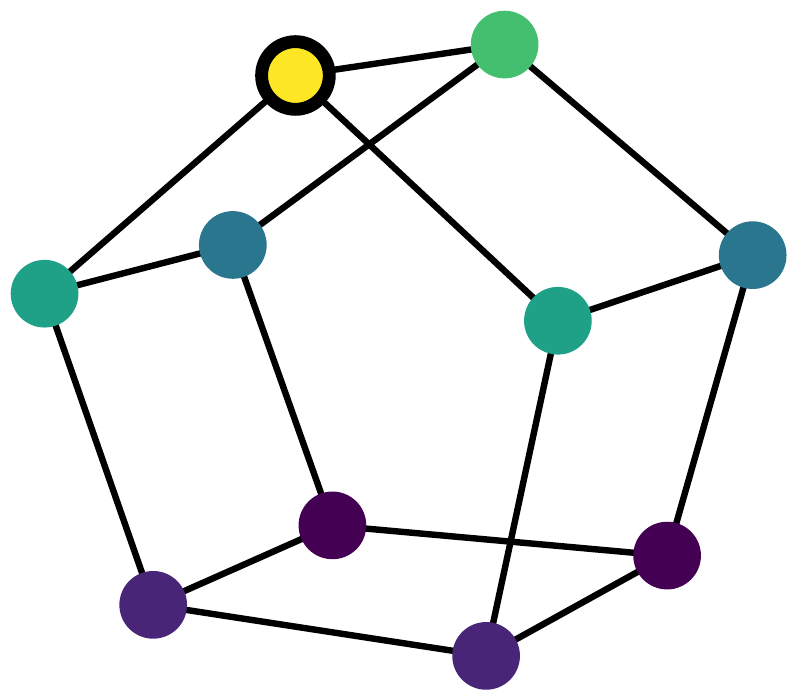}};
\end{tikzpicture}
\caption[]{Left: GeometricKernels-computed samples from a set of Gaussian processes, with covariance given by the heat kernel, on three spaces: the Stanford bunny mesh, the unit sphere $\mathbb{S}_2$, and the hyperbolic space $\mathbb{H}_2$. Right: a GeometricKernels-computed Matérn kernel $k(\mathbin{\tikz\draw[line width={1pt},fill={node_yellow}] (0,0) circle (2pt);},\.)$ on a graph.}
\label{fig:samples}
\end{figure}

The GeometricKernels Python package implements heat and Matérn kernels for a broad class of geometric spaces.
To aid understanding, we review these kernels in \Cref{sec:theory}.
In addition, \Cref{sec:history} includes a brief historical overview of these and related kernel classes.
We now discuss this package's technical capabilities, some of which are illustrated in \Cref{fig:samples}, as well as its design.

\subsection{Classes for Geometric Spaces}

Supported spaces are represented as classes in the \mintinline{python}{spaces} subpackage. Current support includes:
\1 Compact Riemannian manifolds: the unit circle~$\mathbb{S}_1$ via the class \mintinline{python}{Circle}; unit hyperspheres~$\mathbb{S}_n$, $n \geq 2$, via the class \mintinline{python}{Hypersphere}; special orthogonal Lie groups~$\mathrm{SO(n)}$, $n \geq 3$, via the class \mintinline{python}{SpecialOrthogonal}; special unitary Lie groups~$\mathrm{SU}(n)$, $n \geq 2$ via the class \mintinline{python}{SpecialUnitary}.
\2 Non-compact Riemannian manifolds: the hyperbolic spaces $\bb{H}_n$, $n \geq 2$, via the class \mintinline{python}{Hyperbolic}; the manifold $\mathrm{SPD}(n)$, $n \geq 2$, of symmetric positive definite matrices $n \x n$ endowed with the affine-invariant Riemannian metric via the class \mintinline{python}{SymmetricPositiveDefiniteMatrices}.
\3 Meshes, viewed as approximations of smooth two-dimensional surfaces: via the class \mintinline{python}{Mesh}.
\4* Node sets of general undirected graphs with non-negative weights: via the class \mintinline{python}{Graph}. 
\5* Node sets of specialized graphs which are used for representing combinatorial domains: via, for instance, the \mintinline{python}{HypercubeGraph} class.
\6* Edge sets of unweighted undirected graphs, or more generally edge sets of simplicial $2$-complexes: via the \mintinline{python}{GraphEdges} class.
\7* Products of discrete spectrum spaces: among the preceding spaces, this includes everything except non-compact Riemannian manifolds, and is accessed via \mintinline{python}{ProductDiscreteSpectrumSpace}.
\0
A brief discussion of additional spaces one might incorporate in future work is given in \Cref{sec:theory}.

\subsection{Classes for Geometric Kernels}

Although multiple kernel classes technically exist in the library, the class \mintinline{python}{MaternGeometricKernel} in the \mintinline{python}{kernels} subpackage universally covers most use cases, without any need to understand the library internals or the mathematics behind the kernels.
This \mintinline{python}{MaternGeometricKernel} dispatches on the provided space and automatically chooses the internal implementation  of the kernel.
Another useful kernel class is \mintinline{python}{ProductGeometricKernel} which takes a number of geometric kernels $k_i: X_i \x X_i \-> \R$ for $1 \leq i \leq m$---where $X_i$ can be any of the supported spaces, without restrictions---and constructs the product kernel $k: X \x X \-> \R$ on $X = X_1 \x .. \x X_m$ defined by
\[
k((x_1, .., x_m), (x_1', .., x_m'))
=
k_1(x_1, x_1') \cdot .. \cdot k_m(x_m, x_m')
\]
This is more computationally efficient than constructing a \mintinline{python}{MaternGeometricKernel} instance corresponding to a \mintinline{python}{ProductDiscreteSpectrumSpace} and provides additional flexibility, as each of the factors $k_i$ can have its own independent set of hyperparmeters, providing automatic-relevance-determination-like \cite{rasmussen2006} behavior.

\subsection{Approximate Finite-dimensional Feature Maps}

Beyond kernels, the library offers the \emph{approximate finite-dimensional feature maps} corresponding to all the kernels provided.
These are functions $\v\phi : X \-> \R^\ell$ such that the kernel $k$ satisfies
\[
k(x,x') \approx \v\phi(x)^T \v\phi(x)
\]
where $\ell$ is a hyperparameter that controls approximation quality.
Concrete feature map implementations can be found in the \mintinline{python}{feature_maps} subpackage.
For most use cases, an appropriate feature map can be obtained via the function \mintinline{python}{default_feature_map} from the \mintinline{python}{kernels} subpackage.

One can use approximate finite-dimensional feature maps for many purposes.
They are particularly helpful, for instance, to efficiently---that is, without incurring cubic costs---obtain an approximate sample from a Gaussian process $f \~[GP](0, k)$ via
\[
f(x) &= \v\zeta^T \v\phi(x)
&
\v\zeta &\~[N](\v{0},\m{I})
.
\]
We provide convenience utilities for efficiently sampling Gaussian process priors in the \mintinline{python}{sampling} subpackage.
Throughout, care is taken to ensure that computation of the map $x \-> \v\phi(x)$ is done in a fully-automatically-differentiable manner.
For more on feature maps, see the appendices.

\subsection{Multi-backend Design}
GeometricKernels is designed to support multiple numerical computation and automatic-differentiation backends, including \textsc{PyTorch}, \textsc{JAX}, \textsc{TensorFlow}, all with GPU support, as well as a basic \textsc{Numpy} backend for lighter-weight use that does not require automatic-differentiation or GPUs.
This is implemented via multiple dispatch, using function and operator overloading, which is provided by the library \textsc{LAB} \cite{bruinsma2019}.
From the user's viewpoint, one simply needs to include an appropriate import statement, for example
\begin{minted}{python}
import geometrickernels.torch
\end{minted}
to enable PyTorch support.
Alternatively, one can use 
\begin{minted}{python}
import geometrickernels.jax
\end{minted}
to enable JAX support. 
GeometricKernels preserves the array types it is given: as long as all inputs to its functions are \textsc{JAX}-arrays, the output will also be a \textsc{JAX}-array, and similarly for other backends.
All of our computations support batching in the same way as each backend.
Each backend's code works independently: as a result, in principle, all backends can be imported simultaneously---though, we do not do so unless the user explicitly requests it, to avoid bringing in unnecessary dependencies.

\subsection{Integration with Other Packages}

We support integration with Gaussian process packages \textsc{GPyTorch} \cite{gardner2018}, \textsc{GPJax} \cite{pinder2022}, and \textsc{GPflow} \cite{matthews2017}, which are currently the most popular packages in their ecosystems.
This integration is provided via the respective kernel wrapper classes \mintinline{python}{GPyTorchGeometricKernel}, \mintinline{python}{GPJaxGeometricKernel}, \mintinline{python}{GPflowGeometricKernel} in the \mintinline{python}{frontends} subpackage.
We also support standalone use without a frontend, for instance for kernel PCA.

\section{An Illustrative Example}
\label{sec:example}

We now walk through a simple example, which is mirrored on our documentation.
Our goal is to compute the kernel matrix for a Matérn kernel on the sphere.
First, let us import the package using
\begin{minted}{python}
import numpy
import geometric_kernels
from geometric_kernels.spaces import Hypersphere
from geometric_kernels.kernels import MaternGeometricKernel
\end{minted}
where, since we have not imported a backend, the package will use the \textsc{NumPy} backend by default.
To create a kernel one uses
\begin{minted}{python}
hypersphere = Hypersphere(dim=2)
kernel = MaternGeometricKernel(hypersphere)
\end{minted}
which first creates the space---here, a simple two-dimensional unit sphere $\bb{S}_2$---and then, the kernel over the space.
To ensure maximum support for different backends, which might handle trainable parameters differently, we adopt a \textsc{JAX}-inspired design where kernel hyperparameters are stored within a Python dictionary.
This is initialized, and then set, using 
\begin{minted}{python}
params = kernel.init_params()
params["nu"] = np.array([5/2])
params["lengthscale"] = np.array([1.])
\end{minted}
which gives us a parameter object, here called \mintinline{python}{params}.
We now assemble a kernel matrix using three points on the sphere, parameterized by their coordinates in $\R^3$.
For this, we evaluate the kernel via
\begin{minted}{python}
xs = np.array([[0., 0., 1.], 
               [0., 1., 0.], 
               [1., 0., 0.]])
Kxx = kernel.K(params, xs)
\end{minted}
which returns a matrix \mintinline{python}{Kxx}, with whatever array type was supplied within the input variables, namely \mintinline{python}{params}, \mintinline{python}{x1}, and \mintinline{python}{x2}.
We can now use this kernel matrix for whatever we want---for instance, kernel ridge regression---but for illustrative purposes here simply print it.
The obtained values are 
\begin{minted}{python}
print(np.around(Kxx, 2))

# [[1.   0.36 0.36]
# [0.36 1.   0.36]
# [0.36 0.36 1.  ]]
\end{minted}
which, since our inputs are Numpy arrays, produces a Numpy array.
An illustration of how our code works with different backends is shown in \Cref{fig:backends}.

\section{Conclusion}

In this work, we introduced the GeometricKernels Python package.
We documented the kinds of kernels, spaces and other primitives that it supports, and described the key elements of its design philosophy, including seamless GPU and automatic differentiation support within a number of different backends.
We concluded with a simple example illustrating these capabilities of the package.

\section*{Acknowledgments} 

We are grateful to Wessel Bruinsma for his prompt responses and invaluable assistance with the LAB library \cite{bruinsma2019}, which is the foundation of the backend-independence of our package.
Furthermore, we express our gratitude to Thomas Pinder for contributing the first version of the geometric kernel wrapper for his library GPJax \cite{pinder2022}.
While at St. Petersburg State University, PM was supported by the RSF grant N\textsuperscript{\underline{o}}21-11-00047.
IA was supported by the UK Engineering and Physical Sciences Research Council, grant number EP/T517811/1.
NJ is supported by the Wallenberg AI, Autonomous Systems and Software Program (WASP) funded by the Knut and Alice Wallenberg Foundation.
AR is supported by the Accelerate Programme for Scientific Discovery.
AT was supported by Cornell University, jointly via the Center for Data Science for Enterprise and Society, the College of Engineering, and the Ann S. Bowers College of Computing and Information Science. 
VB was supported by an ETH Zürich Postdoctoral Fellowship.

\printbibliography

\appendix

\newpage

\section{Heat and Matérn Kernels on Non-Euclidean Spaces}
\label{sec:theory}

Here we introduce the necessary technical notions, beginning with a brief overview of kernels in general, Gaussian processes, and the specific kernel classes of interest.

\subsection{Brief Review of Kernels}

Let $X$ be a set.
A \emph{kernel} is defined to be a positive semi-definite function $k : X \x X \-> \R$.
By \emph{positive semi-definite function}, we mean that for any $N$ and any collection of data points $\v{x} = (x_1,..,x_N)$, the \emph{kernel matrix} 
\[
\m{K}_{\v{x}\v{x}} = \sbr{k(x_i,x_j) : \begin{gathered}i=1,..,N\\j=1,..,N\end{gathered}}
\]
must be positive semi-definite.
We will be interested in kernels whose domain $X$ is a manifold, mesh, or graph rather than a Euclidean space $\R^n$.

Kernels are a fundamental building block of many machine learning algorithms \cite{scholkopf2002}.
Examples include non-linear classification and regression algorithms, such as for instance kernel ridge regression, as well as kernel-based distance functions such as the maximum mean discrepancy and dimensionality reduction techniques like kernel PCA.
For most of these algorithms, the key computational primitive involves evaluating the kernel function $k(x,x')$ pointwise for some set of data points $x,x' \in X$ chosen by the algorithm.
In addition, certain algorithms of key interest, such as Gaussian process models which quantify uncertainty, often require further computational primitives: to understand what is needed, we introduce them now.

\subsection{Gaussian Processes}

One of the most important classes of kernel-based models today is the class of \emph{Gaussian process} models \cite{rasmussen2006}, which are used for Bayesian learning and decision-making algorithms such as Bayesian optimization \cite{frazier2018,garnett2023} and various forms of active learning.
We introduce these now.

In the simplest case, one considers the non-linear regression model $y_i = f(x_i) + \eps_i$, where $\v\eps\~[N](\v{0},\m\Sigma)$ is the likelihood noise variance, and $f$ is the unknown regression function to be learned.
Gaussian processes are a class of prior distributions on the space of functions $f : X \-> \R$ which are uniquely determined by their mean and covariance kernel, written $f\~[GP](m,k)$.
Such priors are conjugate: letting $m=0$ without loss of generality, the resulting posterior distribution $f \given \v{y}$ is another Gaussian process, with mean and covariance
\[
\mu_{f\given\v{y}}(\.) &= \m{K}_{(\.)\v{x}} (\m{K}_{\v{x}\v{x}} + \m\Sigma)^{-1} \v{y}
&
k_{f\given\v{y}}(\.,\.') &= \m{K}_{(\.,\v{x})} (\m{K}_{\v{x}\v{x}} + \m\Sigma)^{-1} \m{K}_{(\v{x},\.')} 
.
\]
One can also express posterior random functions using \emph{pathwise conditioning} \cite{wilson20,wilson21} as transformations of prior random functions, given by
\[
(f\given\v{y})(\.) = f(\.) + \m{K}_{(\.)\v{x}} (\m{K}_{\v{x}\v{x}} + \m\Sigma)^{-1} (\v{y} - f(\v{x}) - \v\eps)
.
\]
To use these expressions, we require two computational primitives: (1) mirroring general kernel methods, the ability to compute $k(x,x')$, and (2) the ability to sample the prior $f(\.)$ at a potentially-large set of evaluation points jointly, in a computationally-efficient manner with respect to the number of evaluation points.
Of these, (2) generally requires computational primitives beyond pointwise kernel evaluation: a standard approach involves using an \emph{approximate feature map}---that is, a function $\phi: X \-> \R^l$ that $k(x, x') = \innerprod{\phi(x)}{\phi(x')}_{\R^l}$.

\subsection{Heat Kernels}

Principled universal kernels that adapt to the geometry of the input domain are of key importance in the tasks where well-calibrated uncertainty is needed.
In the Euclidean case, the most popular such classes are the squared exponential (RBF, Gaussian, diffusion, heat) and Matérn kernel.
It turns out---see, for instance, \textcite{borovitskiy2020, borovitskiy2021, azangulov2022, azangulov2023}---these classes can be generalized to a number of geometric settings in principled way.
These generalizations can be derived in many different ways: in what follows, we describe them in, arguably, the simplest way possible, along the lines of \textcite{azangulov2022, azangulov2023}.

We begin by introducing the \emph{heat kernel} $k_{\infty,\kappa,\sigma^2}$, which is parameterized by a pair of parameters: the \emph{length scale} $\kappa>0$ and \emph{amplitude} (also called \emph{outputscale}) $\sigma^2$.
This kernel is the infinite-smoothness limit of the more general class of Matérn kernels $k_{\nu,\kappa,\sigma^2}$ which we define in terms of the heat kernel in the sequel.
Let $X$ be a Riemannian manifold, mesh, or graph.
Define 
\[
k_{\infty,\kappa,\sigma^2}(x,x') = \frac{\sigma^2}{C_{\infty, \kappa}} \c{P}(\kappa^2/2,x,x')
\]
where $\c{P}$ is the solution of the differential equation
\[ 
\label{eqn:heat_eqn_problem}
\pd{\c{P}}{t}(t,x,x') &= \lap_{x} \c{P}(t,x,x')
&
\c{P}(0,x,x') &= \delta(x,x'),
\]
where $\Delta_x$ is a suitable notion of the Laplace operator operator on the space of interest, acting on the $x$ variable, and $\delta(x,x')$ is the Dirac delta function, defined appropriately.
The constant $C_{\infty, \kappa}$ is defined to ensure $\int_{X} k_{\infty,\kappa,\sigma^2}(x,x) \d x = \sigma^2$.
On all of the classes of spaces of interest, one can show that~\Cref{eqn:heat_eqn_problem} admits a unique solution within a suitable class of functions.
For our purposes, the key computational question will be how to work with this kernel numerically: we will return to this question momentarily, but first define the full class of Matérn kernels.

\subsection{Matérn Kernels}

Heat kernels are infinitely-differentiable functions: in many applications, such as Bayesian optimization, one would rather work with finitely-differentiable functions because in many cases this has been shown to lead to improved performance.
To do so, we introduce the class of Matérn kernels, defined for $\nu>0$ using the expression
\[ \label{eqn:matern_integral_formula}
k_{\nu, \kappa, \sigma^2}(x,x')
=
\frac{\sigma^2}{C_{\nu, \kappa}}
\int_0^{\infty}
u^{\nu - 1 + n/2}
e^{-\frac{2 \nu}{\kappa^2} u}
\c{P}(u, x, x')
\d u
\]
where $C_{\nu, \kappa}$ is the constant that ensures $\int_{X} k_{\nu,\kappa,\sigma^2}(x,x) \d x = \sigma^2$, and $n$ is a space-dependent constant which is equal to $\dim X$ for manifolds, $0$ for graphs, and $2$ for meshes approximating smooth surfaces.
For the spaces of interest, one can show these integrals are well-defined whenever $\c{P}$ is defined.

In the Euclidean setting, Matérn kernels are usually defined as stationary kernels whose spectral measure is a $t$-distribution, which is a gamma mixture of normals.
Our definition is equivalent to this one in the Euclidean case, but moves the gamma-mixture-of-normals representation into the kernel, rather than spectral, domain---and, as a result, is much more general, as it requires no structure on the space of interest, other than the ability to define a heat kernel.
Heat kernels, in turn, are of fundamental interest in mathematics, and a substantial amount of research has gone into how to define them on a large class of spaces.
This makes our definition particularly simple and convenient.

One can derive the same class of kernels from other viewpoints, as well.
In particular, it is easy to show these kernels coincide with ones arising from the stochastic partial differential equation view of Gaussian processes of \textcite{whittle63, lindgren2011, lindgren2022}, by examining their spectral representations like that of \textcite{borovitskiy2020, borovitskiy2021} and applying the argument in \textcite[Proposition 18]{azangulov2022}.
One can similarly obtain this expression starting from the perspective of reproducing kernels on Sobolev spaces \cite[Appendix D]{borovitskiy2020}, which, as a byproduct, implies the same differentiability properties for these kernels as for their Euclidean counterparts.

From a computational viewpoint, working with Matérn kernels turns out to be similar to working with the heat kernel: as occurs in all cases currently considered in the library, the extra integral can be passed into whatever computational scheme one uses for $\c{P}$, and need not be evaluated numerically.
To understand this better, we now review the class of computational techniques we work with.

\subsection{Fourier Features}

The primary class of computational techniques we implement are based on suitable forms of Fourier analysis to the spaces of interest.
For instance, for \emph{discrete spectrum spaces} like compact Riemannian manifolds, graphs and meshes, these are based off of the Laplacian eigenfunctions---namely, functions $\phi_n$ which satisfy $\lap_x \phi_n = -\lambda_n \phi$.
Using these functions, one can solve the heat equation analytically: after some algebra, obtaining formulas such as
\[
k_{\infty,\kappa,\sigma^2}(x,x') = \frac{\sigma^2}{C_{\infty, \kappa}} \sum_{n=0}^\infty \Phi_{\infty, \kappa}(\lambda_n) f_n(x) f_n(x')
\]
where $\Phi_{\infty, \kappa}(\lambda) = e^{-\frac{\kappa^2}{2}\lambda}$ in the heat kernel case.
By passing the integral through the sum, one can show that Matérn kernels have the same form, with a certain function $\Phi_{\nu, \kappa}$ with an explicit analytic form in place of the function $\Phi_{\infty, \kappa}$.
This expression defines a Mercer-type expansion for the kernel, also known as \emph{manifold Fourier features}.

In our work, we implement a large class of expansions of the kind described above---where the precise expansion used, and precise form of $f_n$, are computed on a case-by-case basis for various manifolds.
For meshes and graphs, computing their analogues---graph Laplacian and mesh Laplacian eigenvectors---is a standard and well-studied problem in numerical linear algebra, with broad use in computer graphics and other areas.

For many spaces, especially for compact Riemannian manifolds, we encounter repeated eigenvalues~$\lambda_n$. 
In such cases, we can sometimes compute the sums  $G_l(x, x') = \sum_{s=1}^{d_l}f_{l, s}(x) f_{l, s}(x')$, where $\cbr{f_{l, s}}_{s=1}^{d_l}$ are some sets of eigenfunctions corresponding to the same eigenvalue $\lambda_l$, without actually computing the individual eigenfunctions $f_{l, s}$. 
This gives a simplified and faster-converging expansion:\footnote{We do not require $\cbr{f_{l, s}}_{s=1}^{d_l}$ to span the whole eigenspace corresponding to a certain eigenvalue in general.}
\[ \label{eqn:kernel_levels}
k_{\nu,\kappa,\sigma^2}(x,x') = \frac{\sigma^2}{C_{\nu, \kappa}} \sum_{l=0}^\infty \Phi_{\nu, \kappa}(\lambda_l) G_l(x, x')
.
\]
For example, the classical addition theorem on the hyperspheres $\bb{S}_n$ identifies the $G_l$, where $\cbr{f_{l, s}}_{s=1}^{d_l}$ are sets of eigenfunctions spanning an entire eigenspace, with Gegenbauer polynomials (see, e.g., \textcite[Appendix B]{borovitskiy2020}).
Another situation like this happens on compact Lie groups.
Here, each eigenfunction $f_n$ can be shown to arise from a unitary irreducible representation of the group.
The functions $G_l$ associated to the sets of eigenfunctions corresponding to the same unitary irreducible representation turn out to be proportional to the character of said representation, something that can be computed exactly and efficiently using abstract algebraic considerations \cite{azangulov2022}.
Importantly, in this case, $G_l$ need not correspond to an entire eigenspace.

GeometricKernels takes advantage of this: we introduce the concept of a \emph{level}, which groups together multiple functions $f_n$ whose associated eigenvalues $\lambda_n$ are the same, but does not necessarily include all eigenfunctions of this kind.
More specifically, levels corresond to the indices $l$ and the corresponding sets of eigenfunctions $\cbr{f_{l, s}}_{s=1}^{d_l}$, as in~\eqref{eqn:kernel_levels} above, depending on the context.

For non-compact Riemannian manifolds---which, in the terminology of our package, are not discrete spectrum spaces---the machinery is more akin to the traditional random Fourier features \cite{rahimi2008}: in this case, the kernels are represented by intractable integrals that can be approximated by Monte Carlo techniques.
Importantly, these approximations---and, indeed, all others available the library---are guaranteed to be positive semi-definite.

\subsection{Future Work}

There are a number of geometric settings which admit kernels that can be computed using methods similar to the ones we consider.
This includes spaces of graphs \cite{borovitskiy2023} and other combinatorial spaces \cite{doumont2025}, vector fields on manifolds \cite{hutchinson2021, robertnicoud2024}, metric graphs \cite{bolin2024}, general cellular complexes \cite{alain2024}, and compact homogeneous spaces \cite{azangulov2022}.
In principle, one could consider kernels for implicit geometry \cite{fichera2023, peach2024, dunson2022, niu2023, he2025}, though supporting this in a backend-independent manner would be considerably more involved.
Incorporating these new settings could increase the package's generality and support new use cases.

\section{Brief Historical Overview of Kernels on Structured Domains}
\label{sec:history}

The development of kernels targeting geometric data began mostly in the the 2000s, during a time when kernel methods such as support vector machines were particularly prominent in machine learning.
The types of data considered include strings~\cite{lodhi2002, leslie2001}, points on statistical manifolds~\cite{lafferty2005}, or even data represented as terms in a formal logic~\cite{gartner2004}.
The most popular setting within this broad area~\cite{ghosh2018}, by a large margin, are kernels \emph{between} graphs~\cite{gartner2003, kashima2003} and kernels \emph{on} graphs---that is, kernels between graph nodes~\cite{kondor2002}.
The interest towards the former extended into the 2010s \cite{vishwanathan2010, yanardag2015, kondor2016} and led to creation of such important tools as Weisfeiler--Lehman kernels~\cite{shervashidze2011} closely connected with graph neural networks~\cite{xu2018}.

The aforementioned graph kernels have been used in many applications.
For example, they have been used to predict mutagenicity, toxicity and anti-cancer activity for small molecules~\cite{swamidass2005}, to predict protein function~\cite{borgwardt2005}, and disease outcome from protein-protein interaction networks~\cite{borgwardt2007}.
Other structured kernels have been used in neuroimaging applications~\cite{jie2013}, analyzing FMRI data~\cite{takerkart2014}, and neural architecture search~\cite{ru2020}.
Many other applications can be found in the review by~\textcite{kriege2020} and books by~\textcite{borgwardt2020, gartner2008}.

Kernels, while constrained to be positive semi-definite, are otherwise a very flexible model class.
Most kernel design research, historically, has focused on maximizing predictive performance in specialized settings, and not on effective uncertainty quantification---a key aspect in Gaussian process regression and Bayesian optimization.
Today, this has changed: due to difficulties with producing meaningful uncertainty using other techniques such as neural networks, designing kernels which produce good uncertainty behavior in practice is a prominent research question.

One line of work, which focuses on good uncertainty quantification \cite{borovitskiy2020, borovitskiy2021, azangulov2022, azangulov2023, yang2024, wyrwal2025}, involves generalizing kernel classes with known good uncertainty behavior in the Euclidean setting, such as the squared exponential (also known as heat, RBF, Gaussian, diffusion) and Matérn kernel classes, to geometric settings.
This class is becoming popular in modern geometric Gaussian process applications \cite{coveney2022, jaquier2022, jaquier2024} and comes with an appealing set of theoretical guarantees \cite{li2023, rosa2023}.\footnote{Notably, the manifold Fourier features and their close relatives associated to Euclidean boundary value problems find impressive applications as general inference techniques \cite{dutordoir2020, solin2020, riutortmayol2023}. \textcite{porcu2024} review diverse applications of Matérn kernels throughout various~disciplines.}
Thus we focus on this class of kernels for our package.
Note that, within it, heat kernels have been classically and extensively studied within the mathematics community \cite{grigoryan2009}, with their variants on certain spaces studied in machine learning, for instance, by \textcite{kondor2002, lafferty2005,gao2019}.

\newpage

\section{Multi-backend example}
\label{fig:backends}

Here, we illustrate an example of how the example of \Cref{sec:example} can be adapted to work using all backends. Note that no actual \textsc{GeometricKernels} code changes, except for an import statement.

\begin{figure}[h!]
\begin{subfigure}{0.49\textwidth}
\begin{minted}{python}
import numpy as np
import torch
import geometric_kernels.torch
from geometric_kernels.spaces import Hypersphere
from geometric_kernels.kernels import MaternGeometricKernel

hypersphere = Hypersphere(dim=2)
kernel = MaternGeometricKernel(hypersphere)

params = kernel.init_params()
params["nu"] = torch.tensor([5/2])
params["lengthscale"] = torch.tensor([1.])

xs = torch.tensor([[0., 0., 1.], 
                  [0., 1., 0.], 
                  [1., 0., 0.]])
Kxx = kernel.K(params, xs)

print(np.around(Kxx.detach().cpu().numpy(), 2))

# [[1.   0.36 0.36]
# [0.36 1.   0.36]
# [0.36 0.36 1.  ]]
\end{minted}
\caption{PyTorch}
\end{subfigure}
\begin{subfigure}{0.49\textwidth}
\begin{minted}{python}
import numpy as np
import jax.numpy as jnp
import geometric_kernels.jax
from geometric_kernels.spaces import Hypersphere
from geometric_kernels.kernels import MaternGeometricKernel

hypersphere = Hypersphere(dim=2)
kernel = MaternGeometricKernel(hypersphere)

params = kernel.init_params()
params["nu"] = jnp.array([5/2])
params["lengthscale"] = jnp.array([1.])

xs = jnp.array([[0., 0., 1.], 
               [0., 1., 0.], 
               [1., 0., 0.]])
Kxx = kernel.K(params, xs)

print(np.around(np.asarray(Kxx), 2))

# [[1.   0.36 0.36]
# [0.36 1.   0.36]
# [0.36 0.36 1.  ]]
\end{minted}
\caption{JAX}
\end{subfigure}
\\[2ex]
\begin{subfigure}{0.49\textwidth}
\begin{minted}{python}
import numpy as np
import tensorflow as tf
import geometric_kernels.tensorflow
from geometric_kernels.spaces import Hypersphere
from geometric_kernels.kernels import MaternGeometricKernel

hypersphere = Hypersphere(dim=2)
kernel = MaternGeometricKernel(hypersphere)

params = kernel.init_params()
params["nu"] = tf.convert_to_tensor([5/2])
params["lengthscale"] = tf.convert_to_tensor([1.])

xs = tf.convert_to_tensor([[0., 0., 1.],
                           [0., 1., 0.],
                           [1., 0., 0.]])
Kxx = kernel.K(params, xs)

print(np.around(Kxx.numpy(), 2))

# [[1.   0.36 0.36]
# [0.36 1.   0.36]
# [0.36 0.36 1.  ]]
\end{minted}
\caption{TensorFlow}
\end{subfigure}
\begin{subfigure}{0.49\textwidth}
\begin{minted}{python}
import numpy as np
# no backend other than numpy is needed
import geometric_kernels
from geometric_kernels.spaces import Hypersphere
from geometric_kernels.kernels import MaternGeometricKernel

hypersphere = Hypersphere(dim=2)
kernel = MaternGeometricKernel(hypersphere)

params = kernel.init_params()
params["nu"] = np.array([5/2])
params["lengthscale"] = np.array([1.])

xs = np.array([[0., 0., 1.],
               [0., 1., 0.],
               [1., 0., 0.]])
Kxx = kernel.K(params, xs)

print(np.around(kernel.K(params, xs), 2))

# [[1.   0.36 0.36]
# [0.36 1.   0.36]
# [0.36 0.36 1.  ]]
\end{minted}
\caption{NumPy}
\end{subfigure}
\end{figure}

\end{document}